\documentclass[runningheads]{llncs}
\usepackage[T1]{fontenc}
\usepackage{graphicx}
\usepackage{verbatim}
\usepackage{subcaption}
\usepackage{multirow}
\usepackage{booktabs}
\usepackage[dvipsnames]{xcolor}
\usepackage{amsmath}
\usepackage[numbers,square]{natbib}
\usepackage{url}     
\usepackage{hyperref}
\usepackage{color}

\begin{document}
\title{Predicting the Progression of Adolescent Idiopathic Scoliosis}
%

\author{Owen Pullen \inst{1} \and
Amir Jamaludin \inst{1} \and
Andrew Zisserman \inst{1}}
\authorrunning{O.\ Pullen et al.}
%
\institute{Visual Geometry Group, University of Oxford, Oxford, UK
\email{\{owen,amirj,az\}@robots.ox.ac.uk}}

\maketitle              
\begin{abstract}
Adolescent Idiopathic Scoliosis is defined as a lateral curvature of the spine that develops during adolescence, without known cause. The condition can result in significant pain and disability, and often progresses rapidly during adolescence. The objective of this paper is to predict the progression of the condition in a temporal sequence from ages 9 to 24, as measured from a sequence of Dual X-ray Absorptiometry (DXA) scans. To this end, we train a transformer model that takes in the curve of the spine to predict curve progression. The model is trained using a large-scale synthetic dataset of spine curves and their time series, covering different curve types and different progression patterns. We show that the model is able to generalise from synthetic to real data by evaluating it on a dataset of real DXA scans covering multiple time points. We find that fine-tuning the model on real data gives a significant boost to performance. The model is able to accurately predict spine curve progression in both scoliosis and normal cases.

\end{abstract}

\section{Introduction}
Scoliosis is a lateral curvature of the spine \cite{taylor_identifying_2013} while Adolescent Idiopathic Scoliosis is Scoliosis that presents in adolescent children without any known cause. Adolescent Scoliosis is more common in women than men and is believed to affect between 0.47 and 5.2\% globally \cite{konieczny_epidemiology_2013}. The condition is no longer screened for in the UK, and is often diagnosed late in adolescence.

\begin{figure}[t]
    \centering
    \begin{subfigure}[b]{\linewidth}
        \centering
        \includegraphics[width=0.775\linewidth]{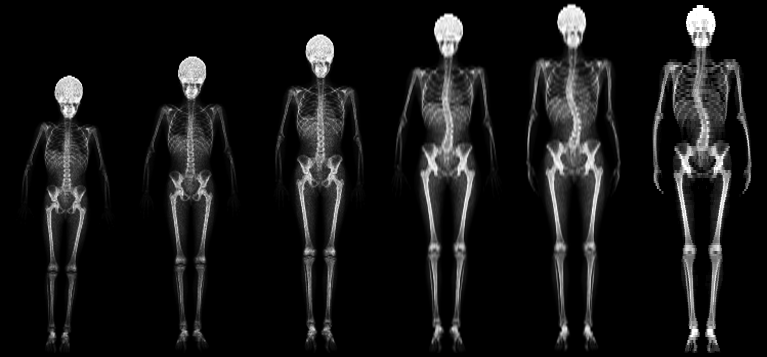}
        \caption{Longitudinal progression of a left lumbar scoliosis curve pattern, during adolescence.}
        \label{fig:progression_1}
    \end{subfigure}

    \vspace{1em}

    \begin{subfigure}[b]{\linewidth}
        \centering
        \includegraphics[width=0.775\linewidth]{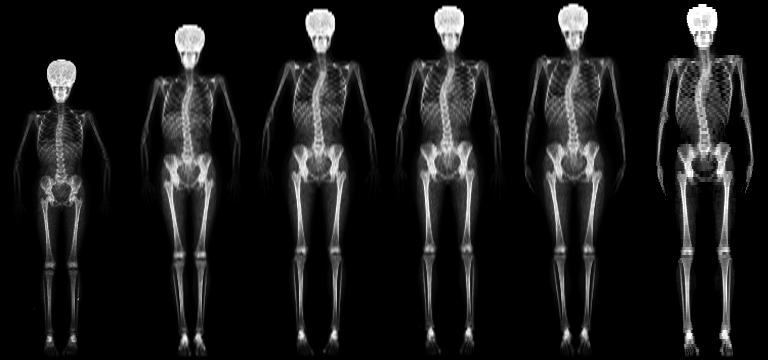}
        \caption{Longitudinal progression of a right thoracic scoliosis curve pattern, throughout adolescence.}
        \label{fig:progression_2}
    \end{subfigure}
    \caption{Longitudinal scoliosis progression in two ALSPAC DXA imaging cohort participants with scans at ages 9, 11, 13, 15, 17, and 24 years. (a) Progression of a left lumbar curve, with development of a secondary compensatory right thoracic curve during late adolescence. (b) Progression from a mild curve at age 9 to a severe right thoracic curve by later follow-up.}
    \label{fig:progression}
\end{figure}

Much attention has been paid to the task of automatically measuring the severity of Scoliosis using medical images \cite{wu2017automatic}\cite{lin_2021_seg}. However, there has been little research attempting to predict the progression of disease. Fig.~\ref{fig:progression} shows two examples of the progression of Adolescent Idiopathic Scoliosis throughout adolescence in the Avon Longitudinal Study of Parents and Children 
(ALSPAC)~\cite{boyd_cohort_2013}, which is the primary dataset used in this paper.

The main objective of this paper is to develop a model that can observe DXA scans at available time points for an individual, and predict the remaining time points. We have two use cases for the model: (1) predicting a single or multiple missing time points in sequences of scans to enable further research into the progression of Scoliosis; (2) predicting the next time point in a sequence of scans so clinicians can determine whether a patient is likely to require more frequent follow up or is at a high risk of rapidly progressing to severe disease status.

As data is scarce, we design a synthetic data pipeline to simulate a variety of changes to spine curves and patterns of temporal progression. This pipeline is used to generate a large-scale training dataset, which is sufficient to train a transformer model. The model is pre-trained on this dataset by masking a random number of samples in the time series and tasking the model to predict these masked samples, conditioned on the observed samples. Following the pre-training on this large-scale synthetic dataset, the model is then fine-tuned on the far smaller real ALSPAC data.
It is evaluated  on a held out test set of unseen real data from the ALSPAC study.
The primary novelty of this work lies in learning to predict the progression of the spinal shape, and thereby enable the prediction of scoliosis severity and clinically relevant phenotypes. Additionally, we designed a synthetic dataset for model pre-training that boosts performance. 

\subsection{Related Work}

\begin{figure}[h!]
    \centering
    \includegraphics[width=0.353\linewidth]{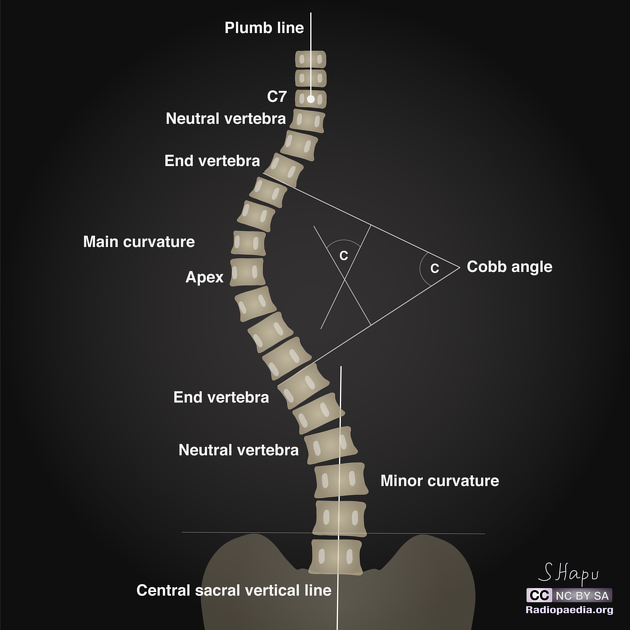}~~~~~~~~
    \includegraphics[width=0.20\linewidth]{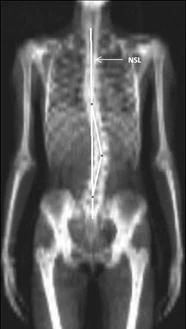}
    \caption{Comparison of scoliosis angle measurements. Left: Cobb angle measurement, where lines are drawn along the endplates of the upper and lower end vertebrae of the curve (figure from Radiopaedia.org~\cite{radiopaedia_cobb}). Right: DXA Scoliosis Method (DSM), which measures a modified Ferguson angle by first defining a ``Normal Spine Line'' to represent a perfectly straight spine. Lines are then drawn from the curve's upper and lower end vertebrae to the curve apex, and the angle at the apex is measured (figure from Taylor et al., 2013~\cite{taylor_identifying_2013}).}
    \label{fig:angle_diagram}
\end{figure}

The Cobb Angle \cite{cobb1948outline} is the most common metric used to measure the severity of Scoliosis. It is obtained from a geometric construction where lines are drawn from the end plates of each vertebrae where the spine curves, and the angle between these lines is the Cobb angle. A visualisation of this is shown in Fig.~\ref{fig:angle_diagram}. It is typically measured on X-rays.

DXA scans have been increasingly used in Scoliosis research and treatment and have been described as being suitable for measuring Scoliosis severity and monitoring progression \cite{taylor_identifying_2013}. The DXA Scoliosis Method (DSM) is similar to the Cobb angle, but adapted for DXA scans where the end-plates of vertebrae are not always clearly visible. Instead of using the end-plates, a line is drawn to represent a perfectly straight spine. Lines are drawn from the first and last vertebrae of the Scoliotic curvature to the curve apex, and the DSM angle is measured between them. Taylor et al. proposed lower threshold of 6\textdegree~to diagnose Scoliosis from DXA scans \cite{taylor_identifying_2013}. DXA scans are taken supine and this reduces the spinal curvature and DSM angle.

There are multiple examples of measuring Cobb angle in X-Rays that achieve excellent results with high accuracy. An example is the  Seg4Reg 
method~\cite{lin_2020_seg}, winner of MICCAI AASCE spinal curvature estimation challenge~\cite{miccai_aasce_2019}, and several other examples~\cite{wu2017automatic}, \cite{lin_2021_seg}, \cite{ZHANG2022sco}. On DXA scans, there exist several methods to compute the DSM ange including:  regressing the angle from the image and segmentation masks~\cite{jamaludin_predicting_2019}; measuring the curvature from the spine segmentation and then transforming this into the DSM angle~\cite{jamaludin2023predicting}; and measuring the DSM angle directly by attempting to automate the DSM process by constructing triangles on the curves and then calculating the angle~\cite{bourigault2022scoliosis}.

Attempts to determine future Cobb angle as a regression task using classical machine learning methods. Other attempts to apply more modern neural network architectures to predict progression of Scoliosis, have focused on predicting severe disease as a binary outcome \cite{Li2024-cc}. 
\section{Dataset: the ALSPAC study}
\begin{figure}[t]
    \centering
    \begin{subfigure}[b]{0.50\textwidth}
        \centering
        \includegraphics[width=\linewidth]{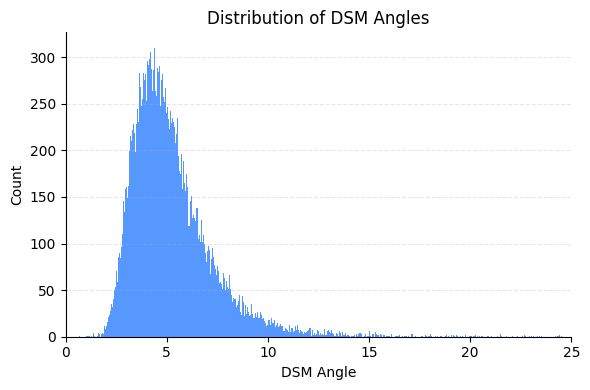}
        \caption{Distribution of DSM and Cobb angles in the ALSPAC dataset. Larger angles are uncommon but represent more severe scoliosis. The x-axis is clipped at 25\textdegree, although DSM angles greater than 25\textdegree are present in the dataset.}
        \label{fig:angle_hist}
    \end{subfigure}
    \hfill
    \begin{subfigure}[b]{0.48\textwidth}
        \centering
        \includegraphics[width=\linewidth]{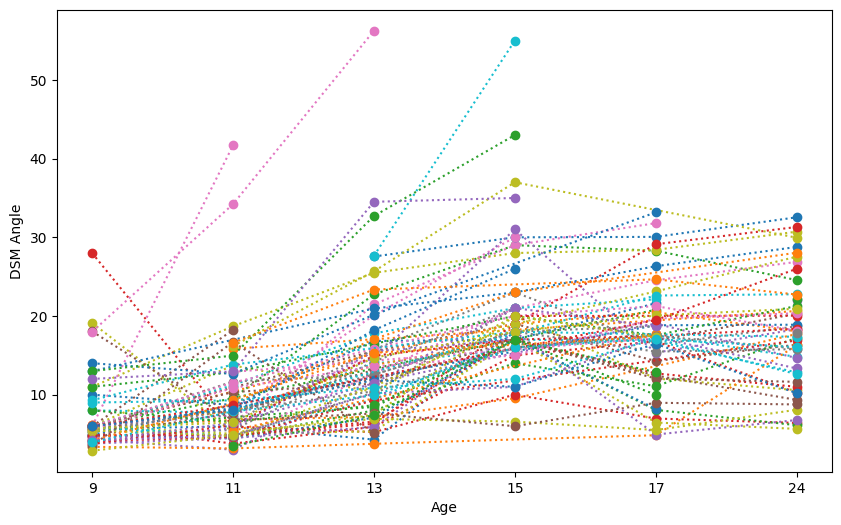}
        \caption{Longitudinal angle sequences can be visualised by connecting measurements across timepoints, illustrating progression patterns within clusters of individuals. AIS cases with angles above 16\textdegree{} are shown for illustrative purposes.}
        \label{fig:sequences_cluster}
    \end{subfigure}
    \caption{Overview of scoliosis angle distributions and progression sequences.}
    \label{fig:combined}
\end{figure}

The Avon Longitudinal Study of Parents and Children (ALSPAC) is a transgenerational cohort study that follows parents and their children over time \cite{boyd_cohort_2013}. Within the ALSPAC study, adolescent participants were scanned at ages 9, 11, 13, 15, 17, and 24 using a DXA scanner, resulting in total-body DXA scans.
Table~\ref{tab:alspac_counts} shows the summary statistics of the dataset, and
Fig.~\ref{fig:angle_hist} shows the distribution of angles. As DXA scans are taken supine the DSM angle is less than a typical Cobb angle from an X-ray. For this work, a lower threshold of 16° was used for the onset of severe disease status. We believe this is equivalent to the commonly used 25° X-Ray, Cobb angle based, threshold used for disease which requires increased monitoring and possible bracing \cite{Li2024-cc}.

\begin{table}[h]
    \centering
    \begin{tabular*}{0.75\textwidth}{l|@{\extracolsep{\fill}}rrrrrrr}
        \hline
         & 9 & 11 & 13 & 15 & 17 & 24 & Total\\
         \hline
         Number of Scans & 7167 & 7001 & 6038 & 4591 & 4823 & 3853 & 33,473 \\
         \hline
    \end{tabular*}

    \vspace{0.5em}

    \begin{tabular*}{0.75\textwidth}{l|@{\extracolsep{\fill}}rrrrrr}
         \multicolumn{7}{c}{Missing Time Points} \\
         \hline
            & 0 & 1 & 2 & 3 & 4 & 5\\
            \hline
         Number of Missing & 1820 & 1800 & 1469 & 1202 & 1241 & 1589\\
         \hline 
    \end{tabular*}
   \caption{ALSPAC scan distribution for 9,121 individuals across age timepoints. The top panel shows the number of participants scanned at ages 9, 11, 13, 15, 17, and 24 years. The bottom panel shows the number of individuals with N missing timepoints due to missed appointments or loss to follow-up.}
    \label{tab:alspac_counts}
\end{table}

Sequences can be visualised to show the overall direction of progression in individuals, see Fig. \ref{fig:sequences_cluster}. In individuals with a scan above 16\textdegree~most individuals increase gradually or do not progress to severe disease rapidly, however, a small minority of individuals progress rapidly to severe disease.

\subsubsection{Real evaluation data.}
The ALSPAC dataset contains DXA scans from 9121 individuals. For the evaluation of the model, a randomly sampled subset of 50\% of the 9121 individuals, obtaining 4561 individuals with 16,708 DXA scans, ensuring that this subset is well-stratified by severity. 8 participants that had surgical interventions have been censored across all sets, right censoring was used at the timepoint surgery was first recognised. It is possible that some individuals may have been treated with non-surgical bracing during their follow-up period, however we do not have access to this information. 
\section{Learning to Predict Progression }

To predict the progression of Adolescent Idiopathic Scoliosis (AIS), we employ an encoder-only transformer. Given that the task involves measuring the progression of angles and is relatively simple, we find it feasible to generate synthetic data that mimics real distributions. The model is pre-trained on this synthetic data and evaluated zero-shot and then fine-tuned on real-data data and evaluated on a held out test set of real ALSPAC data.

\subsection{Synthetic Data Creation}
As transformer models are prone to over fitting and the dataset of real examples is small, a large scale synthetic data is generated to train the model. The creation of the synthetic dataset is guided by the real dataset, as bounds for each curve type were observed and points were uniformly sampled from within this range to create a synthetic training dataset.

In the following, we describe the pipeline for generating synthetic data, which consists of sequences of spine curves. All generated curves have the same vertical extent. The real data are normalised to this same vertical length prior to input to the trained model.

First, the location of the scoliosis is sampled from the following: (1) Lumbar, (2) Thoracic, (3) Thoracolumbar, or (4) Normal examples with no significant curvature. Then, if positive for scoliosis, a left and right curve pattern was created, two examples for each region were created a single curve and a second type with compensatory curve, also known as an S-shaped curve.

To create a time series. A pre-defined curve is taken, $T_0$, a total rate of change is then randomly sampled from a uniform distribution and broken down into factors. The control points at the apex of the curve at $T_0$ are then multiplied by their factors $F_0$ to create $T_1$. At $T_1$ the process is repeated until $T_6$, the final point in the time series is created.

To make each curve type to as realistic as possible the minimum and maximum values of the peaks for each curve type were taken from our real data and the peaks of the curves in the synthetic data was uniformly sampled within this range to provide a series of valid peaks for each curve type. This creates the x-displacement to create a 1-d curve corresponding to a specific Scoliosis phenotype. We do not believe that the distributions of synthetic and real data need to match, as this would recreate the real data's issues with severe Scoliosis data sparsity in the synthetic data.

\begin{figure}[t]
    \centering
    \includegraphics[width=0.5\linewidth]{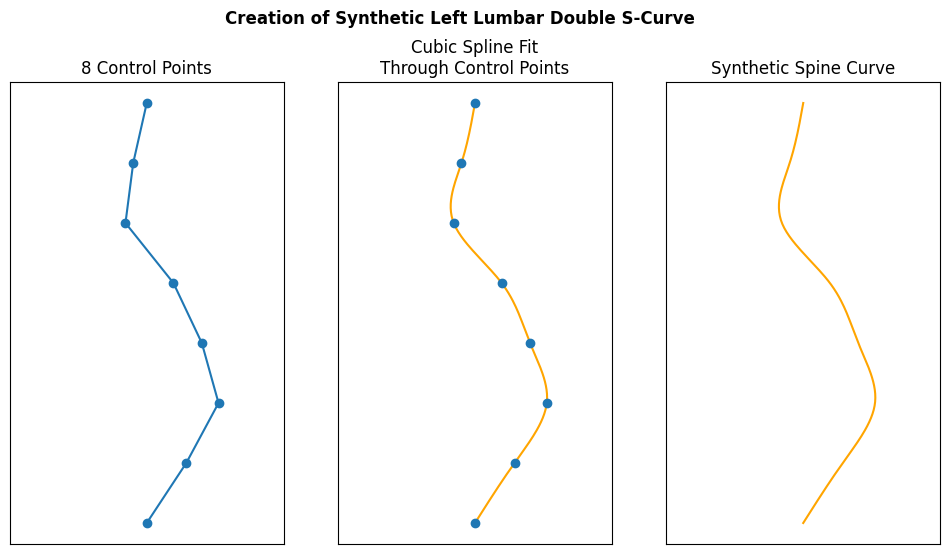}
    \caption{Control points used to create a Double (S-shaped) synthetic curve with a primary peak in the left lumbar region (direction [left] is defined from the perspective of the participant). The synthetic curve is obtained from an interpolating cubic spline fitted through the control points.}
    \label{fig:dsets:ctrl_spline}
\end{figure}

To design each synthetic curve phenotype different subsets of eight control points were moved along the x-axis only. By moving different control points it is possible to create 1-dimensional synthetic examples of different Scoliosis phenotypes. The creation of a synthetic Left Lumbar S-curve from control points is shown in Fig.~\ref{fig:dsets:ctrl_spline}.

\begin{figure}[t]
\begin{subfigure}{\linewidth}
    \centering
    \includegraphics[width=0.9\linewidth]{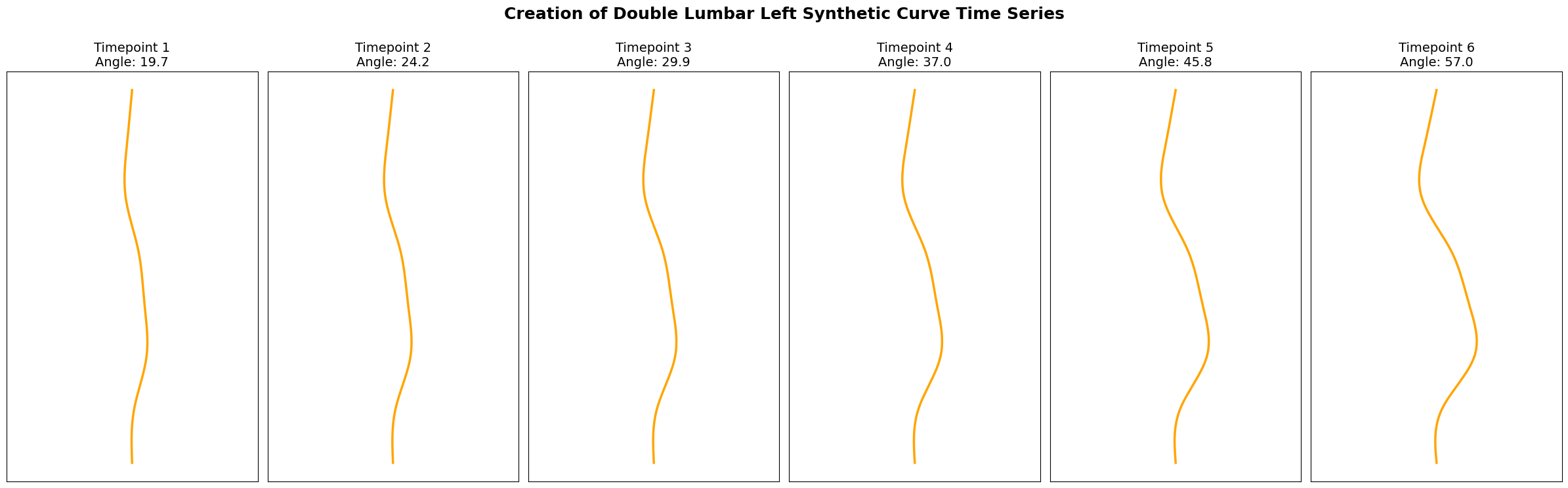}
\end{subfigure}

\begin{subfigure}{\linewidth}
    \centering
    \includegraphics[width=0.9\linewidth]{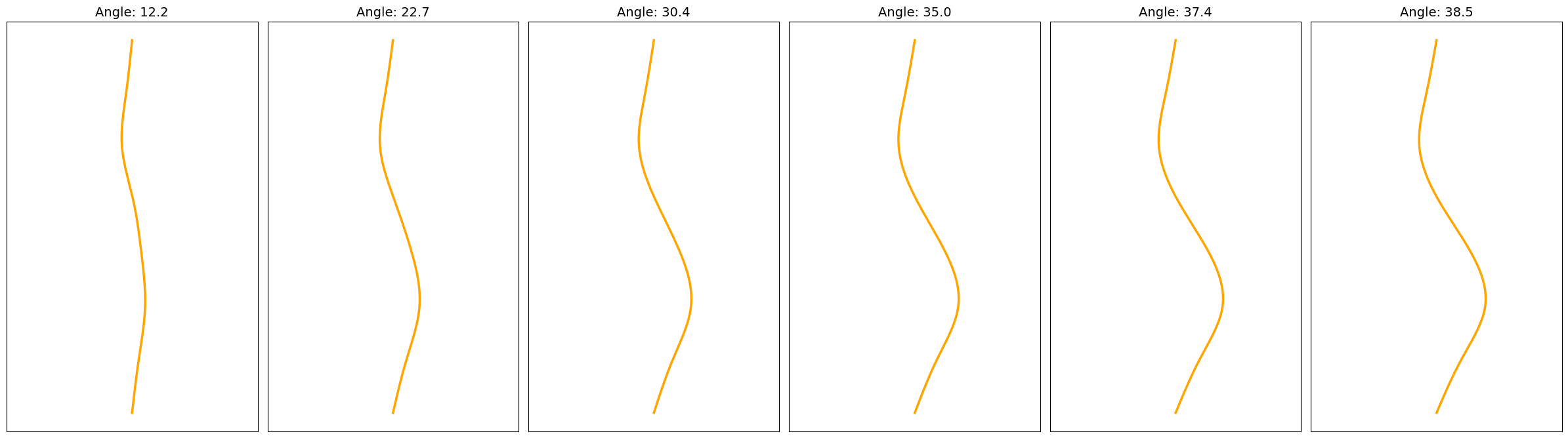}
\end{subfigure}
\caption{Example synthetic curve sequences. The top row shows a continually increasing S-shaped scoliosis case, with the primary curve located in the left lumbar region and increasing throughout the full time series. The bottom row shows a scoliosis case with onset in the left lumbar region, followed by an increasing pattern that plateaus midway through the sequence. Curves are generated by shifting the peak control points along the x-axis, corresponding to lateral spine displacement, and fitting a cubic spline through the control points at each time point.}
\label{fig:dll_ctrl_increase}
\end{figure}

This curve is used as a starting point for the time series of six time steps. The curve's control points were moved in the x-axis to model progression, stabilisation or a reduction in severity of Scoliosis. Four broad progression classes were created: (1) increasing, (2) plateauing followed by increasing, (3) stable, and (4) stable decrease. At each time step the control points are moved in the x-axis to change the x displacement of the curve to create an increasing or other pattern. Overall minimum and maximum increases were captured from the ALSPAC (real) dataset. The multiplicative values for the overall change in x were uniformly sampled from a distribution between these minimum and maximum values. These are distributed among the control points as factors where their product is equal to multiplying by the total value. An example of the control points being manipulated to increase the size of a left lumbar S-shape curve is shown in 
Fig.~\ref{fig:dll_ctrl_increase}.

\subsubsection{Distributions.}
Each synthetic of the 6 timepoints has 693,289 observations. Table~\ref{tab:synthetic_n} shows counts of each of the different curve types and change patterns created in the synthetic dataset.

\begin{table}[h!]
    \centering
\begin{tabular}{llrrrrrr}
\toprule
 & Change & \multicolumn{2}{r}{Lumbar} & \multicolumn{2}{r}{Thoracic} & \multicolumn{2}{r}{Thoracolumbar} \\
 & Pattern & Left & Right & Left & Right & Left & Right \\
 &  &  &  &  &  &  &  \\
\midrule
\multirow[t]{4}{*}{Single} & Increasing & 95550 & 91620 & 92856 & 81408 & 77382 & 67818 \\
 & Increasing Plateau & 316656 & 301950 & 307128 & 263958 & 255438 & 221142 \\
 & Stable & 60000 & 60000 & 60000 & 60000 & 60000 & 59994 \\
 & Stable Decrease & 60000 & 60000 & 60000 & 60000 & 60000 & 59988 \\
\cline{1-8}
\multirow[t]{4}{*}{Double} & Increasing & 79842 & 80598 & 82116 & 78144 & - & - \\
 & Increasing Plateau & 257028 & 258456 & 263682 & 248730 & - & - \\
 & Stable & 60000 & 60000 & 60000 & 60000 & - & - \\
 & Stable Decrease & 60000 & 60000 & 60000 & 60000 & - & - \\
\cline{1-8}
\bottomrule
\end{tabular}
    \caption{Counts for each synthetic data class, aggregated across subclasses. Increasing curves include both gradual and delayed increases in severity, while increasing plateau curves reach their peak at different timepoints within the sequence. An additional 480,000 normal curves follow the stable pattern, with no meaningful increase in disease severity or status over time.}
    \label{tab:synthetic_n}
\end{table}

\subsubsection{Split.}
For the synthetic data we use a 80:10:10 train/val/test split. Each subclass was split individually in this way and then concatenated into train, validation and test datasets. This ensured class balance between train, validation and test.

\subsection{Curve Progression Model}
\begin{figure}[h]
\begin{subfigure}[b]{0.48\textwidth}
    \centering
    \includegraphics[width=5cm]{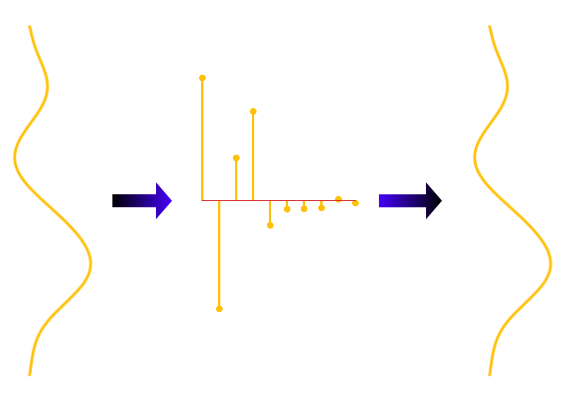}
    \caption{The spine curve be easily decomposed into its sine series coefficients using Fourier analysis, here, using a discrete sine transform. The original signal can be reconstructed by applying an inverse of the transform and summing the resulting sine curves.}
    \label{fig:fourier_fig}
\end{subfigure}
\hfill
\begin{subfigure}[b]{0.48\textwidth}
    \centering
    \includegraphics[width=4.5cm]{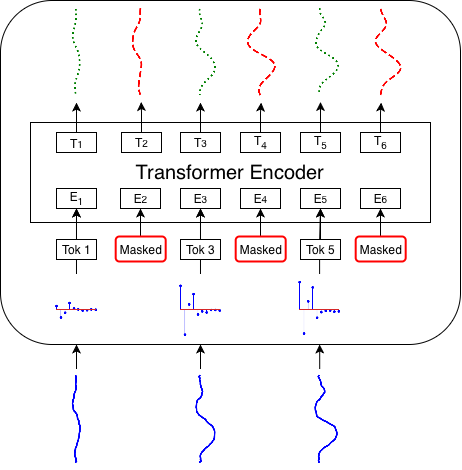}
    \caption{The encoder transformer architecture used to predict missing curves (corresponding to the masked entries). The curves measured from the DXA scans are represented by a low dimensional Fourier expansion, and then mapped to the input dimension of the transformer.}
    \label{fig:model}
\end{subfigure}
\caption{(a) Fourier transform to sine coefficients and reconstruction from coefficients. (b) Diagram of transformer encoder architecture predicting masked coefficients.}
\end{figure}

A lightweight encoder only transformer, based on the BERT  
architecture~\cite{devlin2018bert}, is used. The model ingests a sequence of six tokens, each representing the spine curve at one time point for an individual or a masked token if that scan is missing, and outputs a token representing the spine curve at all time points (i.e.\ including any that are masked). Position encoding is used to represent the time points, and is added to the input vectors to the transformer representing the curves. The masking tokens are learnt. The model consists of 2 transformer layers with 2 attention heads, the input dimension is $d_{model} = 128$, and the feedforward dimension is $d_{ff} = 256$. See Fig.~\ref{fig:model}.

\subsubsection{Curve extraction \& Curve Representation.}
A spine is segmented from  a DXA scan using the segmentation network from~\cite{jamaludin_predicting_2019}. The curve is then obtained from the midpoints of the spinal segmentation mask, producing a 2-dimensional curve.
This is  rotated so that the start and end points are vertically above each other. The curve is then zero bounded and length normalized, allowing it to be represented by a compact 10 dimensional Fourier sine expansion using the method of~\cite{Pullen25}. 
An example of how a spine curve can be transformed into its component sines and reconstructed is shown in Fig.~\ref{fig:fourier_fig}. We use this representation to provide a simplified low dimensional and slightly smoothed representation of the spine. The low-dimensional Fourier representation is used as the input to the transformer.

\noindent {\bf Transformer input/output.}
The model input is a real-valued 10 coefficient sine series, see Fig.~\ref{fig:fourier_fig}. This is projected up to the transformer model dimension of $d_{model} = 128$ for input to the model. The output of the transformer model is projected back to 10 sine coefficients using a linear layer. The sine curve can then be reconstructed.

\subsection{Training}
The model was pre-trained on synthetic data to reconstruct randomly masked timepoints, masked using a learned mask token, for 500 epochs, with the best preforming checkpoint selected for evaluation. An L1 loss was used with the AdamW optimiser with an initial learning rate of $1e-4$ and a weight decay of $1e-4$. Dropout was set to $p = 0.1$. The training reconstruction loss was self-supervised. The loss was computed on the difference between the 10 sine coefficient predictions and ground truth. There are 693,289 synthetic time series that each simulate an individual in the train set, each synthetic time series had 6 time points for a total of 4,159,734 simulated time points.

The model was fine-tuned on ALSPAC using a 45:5:50 train:validation:test split, the data is split by individual so that no individual can appear across multiple sets (i.e. an individual in train will not be in test). To ensure comparability the same test set is used to evaluate the pre-trained model and after fine-tuning. To fine tune the model two losses are constructed. A similar self-supervised coefficient loss is used, although during fine-tuning an L2 loss is used and the unmasked timepoints are down weighted to ensure the loss is primarily focused on masked reconstruction on real sequences. A second L2 loss is constructed between the predicted and measured ground truth angles. These two losses are re-scaled and summed before back-propagation.

\subsubsection{Augmentations.}
The models were trained using a 1 dimensional Gaussian noise augmentation during training. Gaussian augmentation added high frequency noise the the curve before transformation into the Fourier domain. A 1-d dimensional Gaussian noise augmentation is also used during fine-tuning to reduce over-fitting.

\subsection{DSM Angle Regression Network} 
To measure the DSM angle we use an angle regression network. The network inputs the low-dimensional Fourier representation (that are used as the input and output of the transformer encoder model). In~\cite{Pullen25}, it was shown that the  low-dimensional Fourier representation can be fed into a small MLP to reliably and accurately estimate Scoliosis and the associated phenotypes. Therefore we have designed a similar network to regress the DSM angle from the Fourier representation.

The Angle Regression network is a minimally adapted version of the network from~\cite{Pullen25}, consisting of a Multi-Layer Perceptron (MLP) with an input size of 10 to match the Fourier representation, and 5 linear layers of hidden width 64 with ReLU non-linear activation functions between each layer. The model is fitted with a regression head instead of classification heads. 

\begin{figure}[h!]
    \centering
    \includegraphics[width=0.40\linewidth]{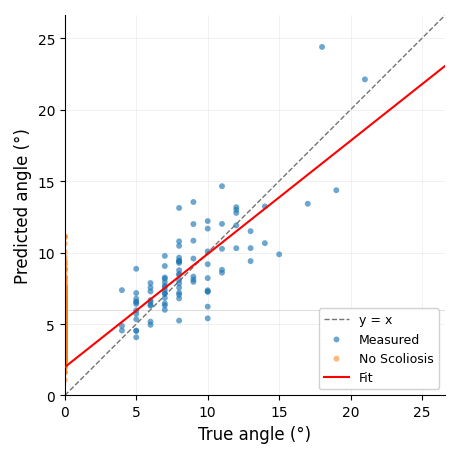}
   \caption{Evaluation of the angle regression network. The network showed excellent performance on real-world ground-truth angles annotated by expert clinicians specialising in scoliosis management and treatment. Left: normal, non-scoliotic cases with ground-truth DSM angles $<6$\textdegree{} (orange). Measured angles (blue).}
    \label{fig:reg_eval}
\end{figure}

The angle regression network was trained on annotated ALSPAC and UK Biobank data and evaluated on the ALSPAC dataset. An evaluation on ALSPAC data previously unseen by the network is shown in Fig.~\ref{fig:reg_eval}. The regression network exhibits excellent results when tested on real-world data with a test MAE, of 1.62\textdegree~ and Mean Absolute Percentage Error (MAPE) of 19.28\% and a Pearson's correlation coefficient of 0.80, which is comparable to \cite{jamaludin2023predicting} which reports a Pearson's of 0.82 with an MAE of 1.9\textdegree. Due to having limited annotated ground truth, the regression network is trained on a separate 80:10:10 split of the ALSPAC data, using additional UK Biobank data in the train set.

\section{Results}
The curve progression model is tested on previously unseen testing data reserved from the synthetic dataset and also on held out test set of the real ALSPAC dataset. In the synthetic data 10\% of the data was reserved for testing.

Both the synthetic and real data is comprised of six scans. The synthetic data is created with no missing time points. However, as the ALSPAC data is from a real-world cohort study, participants frequently missed appointments so the majority of individuals have less than six time points. This increases the difficulty of the task as it obscures important information, missing time points in different areas can also affect the difficulty of the task, i.e.\ a single missing time point immediately preceding the predicted time point could impede next time point prediction much more than a missing time point earlier in the sequence.

Performance has been tested on two tasks: (1) Interpolating a single masked time point at age 15, where all other time points are visible;  (2) Extrapolating to predict the next time point at ages 15, 17 and 24, where all future time points are masked, to prevent the model of from looking ahead, e.g.\ predicting masked time at age 15 from unmasked time points at ages 9, 11 \& 13, where time points at age 17 \& 24 are also masked.

\paragraph{Evaluation metrics.}
The metric used to measure the success of reconstruction tasks is the proportion of predictions within two, three, four and five degrees of the target angle. This is a threshold accuracy metric, where the accuracy shows the proportion of  predictions within $n$\textdegree~of the target.

\subsection{Baselines}
For all experiments we construct simple baselines, these fully reconstruct the curve and allow all curves the angle to be measured by the same frozen DSM angle regression network. For both experiments we construct a baseline in the curve space and in PCA space. PCA space baselines are reconstructed into curves before angles are measured.

\noindent \textit{Interpolating the curves.} The masked curve is reconstructed using an inverse age-distance linear interpolation of the observed curves, a multiplier is used to represent distance from the target in age, ensuring curves closer to the target contribute more to the reconstruction:
$
\mathbf{x_{t^{*}}} = \sum_{i} \tilde{\lambda}_i\, \mathbf{x}_{t_i},
\tilde{\lambda}_i = \frac{\lambda_i}{\sum \lambda_{i ...n}},
\lambda_i = \frac{1}{ \lvert \mathrm{age}_{t^{*}} - \mathrm{age}_{t_i} \rvert}
$ where $x_{t_i}$ is the observed curve at age $t_i$, $t^{*}$ is the masked timepoint, and the sum runs over all observed timepoints. This is done in both curve and PCA space to provide two interpolation baselines.

\noindent \textit{Extrapolating the curves.}
An SVD is used to reduce the dimensionality of the observed curves of a sequence: either raw coordinates in curves space or projected PCA space coefficients. A linear regression is then fitted to the remaining points, yielding a centroid c and unit direction v. Each observed point is projected to give its position along v. A per-year rate is estimated by linearly regressing these positions against age. The forecast is then obtained by sampling the at the next age-timepoint: stepping along v from the fitted positions by the per-year rate times the age gap. This is computed in both curve space and PCA space to provide two baselines.

\noindent \textit{Principal component analysis (PCA).}
The curves were represented in a principal-component basis learned from all timepoints of the ALSPAC (real data) training curves, with $K=10$ components. The first $K=10$ components retain $99.5\%$ of the variance. After this, the PCA basis was applied to the test set for evaluation. The observed curves of a sequence are projected into the principal-component subspace. 

\noindent \textit{Training Only on ALSPAC Data.} To determine if pre-training on the synthetic
dataset aides performance we trained the model only on the ALSPAC dataset.

\subsection{Interpolation results}
\begin{table}[h!]
    \centering
    \begin{tabular}{ll|cccc}
    \hline
    \multicolumn{5}{c}{Interpolation - Age 15 Masked} \\
    \hline
    \multirow{2}{*}{\shortstack{Severity\\(DSM - Degrees)}} & \multirow{2}{*}{Method} & \multicolumn{4}{c}{Masked Time Point(\%)} \\
         && 2\textdegree & 3\textdegree & 4\textdegree & 5\textdegree \\
         \hline
         \multirow{5}{*}{0-6} & Linear Interpolation & 91.95 & \textbf{99.44} & \textbf{100.00} & \textbf{100.00} \\
         & PCA & 91.57 & \textbf{99.44} & \textbf{100.00} & \textbf{100.00} \\
         & ALSPAC Train & \textbf{92.53} & 98.95 & 99.82 & 99.91 \\
         & Pre-train & 89.37 & 97.36 & 99.38 & 99.74 \\
         & ALSPAC Fine-tune & 91.21 & 98.68 & 99.65 & 99.74 \\
         \hline
          \multirow{5}{*}{6-10} & Linear Interpolation & 42.99 & 72.43 & 92.52 & 97.66 \\
         & PCA & 42.52 & 74.77 & 92.52 & 97.20 \\
         & ALSPAC Train & 63.06 & 82.88 & 95.50 & \textbf{100.00} \\
         & Pre-train & 58.56 & 78.38 & 94.59 & 99.10 \\
         & ALSPAC Fine-Tune & \textbf{68.47} & \textbf{90.99} & \textbf{96.40} & \textbf{100.00} \\
         \hline
         \multirow{5}{*}{10-16} & Linear Interpolation & 28.12 & 46.88 & 65.62 & 68.75 \\
         & PCA & 28.12 & 46.88 & 65.62 & 68.75 \\
         & ALSPAC Train & 58.18 & 76.36 & \textbf{89.09} & \textbf{96.36} \\
         & Pre-train & \textbf{56.36} & 74.55 & 81.82 & 90.91 \\
         & ALSPAC Fine-tune & 50.91 & \textbf{78.18} & 85.45 & 94.55 \\
         \hline
         \multirow{5}{*}{16+} & Linear Interpolation & 0.00 & 12.50 & 50.00 & 50.00 \\
         & PCA & 0.00 & 25.00 & 50.00 & 62.50 \\
         & ALSPAC Train & 38.89 & \textbf{83.33} & \textbf{94.44} & \textbf{100.0} \\
         & Pre-train & 44.44 & 72.22 & 83.33 & 94.44 \\
         & ALSPAC Fine-Tune & \textbf{55.56} & \textbf{83.33} & \textbf{94.44} & 94.44 \\
         \hline
    \end{tabular}
    \caption{{\bf Interpolation.} Predicting masked time points within the sequences of scans. Methods: Linear Interpolation, PCA Interpolation (PCA), trained on ALSPAC data only (ALSPAC Train), pre-trained model (Pre-train), and fine-tuned on ALSPAC (Fine-Tune). Everything is evaluated on the ALSPAC test set.}
    \label{tab:res_masked_tf3}
\end{table}

\begin{figure}[h!]
    \centering
    \includegraphics[width=\linewidth]{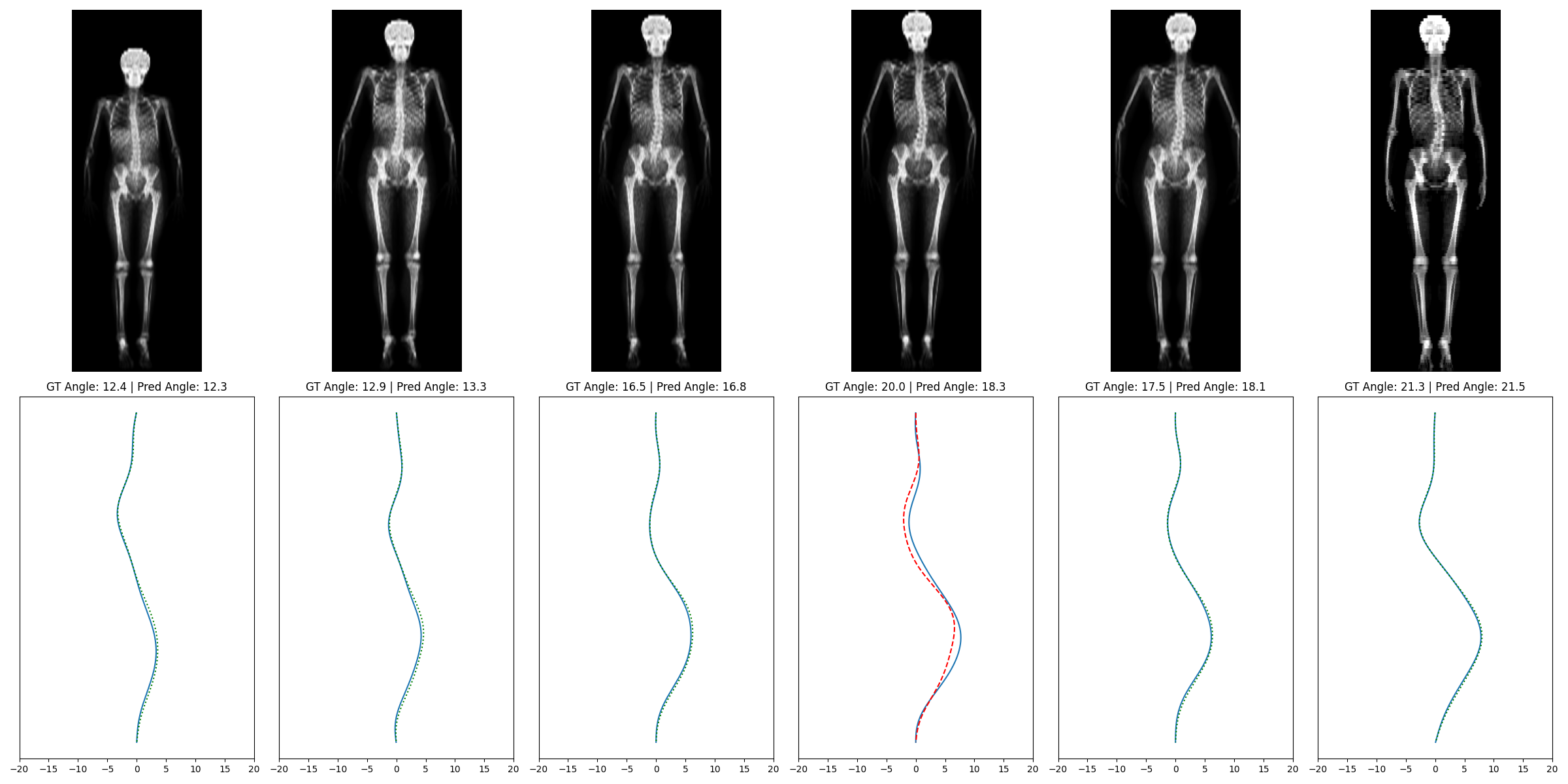}
    \caption{Single masked time point prediction with 4 unmasked time points at 11, 13 17 and 24 (green dotted) to predict masked time point 15 (red dashed). The scan at age 9 is missing but the network is still able to predict accurately.}
    \label{fig:single_example_1}
\end{figure}

The model's ability to predict missing time points within a sequence of six scans has been tested by the age 15 masked time point. The other timepoints in the time series are unmasked and 1 is masked and is predicted. The results of predictions are shown for errors within 2\textdegree, 3\textdegree, 4\textdegree~\&~5\textdegree~in Table \ref{tab:res_masked_tf3}.

The model shows good results on interpolating a single masked time point at age 15, the angle is predicted within 5\textdegree~in the vast majority of cases across and with high accuracy at 2\textdegree, 3\textdegree, and 4\textdegree. The baselines outperform the models in the non-case category as there is minimal change in the 0-6\textdegree~non-scoliosis participants, so the baselines excel at predicting where there is little or no change across the timeseries of scans. Participants with smaller DSM angles are less likely to progress to severe disease, so represent an easier task. Across all other disease severities (all patients who have Scoliosis). The model trained on ALSPAC data only, does well across most interpolation tasks often out-preforming the simple baselines and has comparable results with the models pre-trained on synthetic data. The results beat the baselines for the interpolation task, and fine-tuning the model improves performance compared to the model that is pre-trained (only) across both the 2\textdegree~threshold and the 5\textdegree~threshold, showing fine-tuning makes the model more accurate and more consistent. An example of a interpolation prediction is shown in Fig.~\ref{fig:single_example_1}.

\subsection{Extrapolation results}
\begin{table}[h!]
    \centering
    \setlength{\tabcolsep}{4pt}
    \begin{tabular}{@{}llcccccccc@{}}
        \toprule
        & & \multicolumn{2}{c}{0--6} & \multicolumn{2}{c}{6--10} & \multicolumn{2}{c}{10--16} & \multicolumn{2}{c}{16+} \\
        \cmidrule(lr){3-4}\cmidrule(lr){5-6}\cmidrule(lr){7-8}\cmidrule(lr){9-10}
        Age & Method & 2\textdegree & 3\textdegree & 2\textdegree & 3\textdegree & 3\textdegree & 4\textdegree & 4\textdegree & 5\textdegree \\
        \midrule
        \multirow{4}{*}{15} & Linear & 70.92 & 85.13 & 53.23 & 74.63 & 38.46 & 50.00 & 62.50 & 75.00 \\
         & PCA & 70.64 & 84.94 & 53.23 & 75.12 & 38.46 & 46.15 & 50.00 & 75.00 \\
         & ALSPAC & \textbf{94.18} & \textbf{98.76} & 45.27 & 74.63 & 38.46 & 50.00 & 50.00 & 75.00 \\
         & Pre & 81.91 & 92.44 & 48.54 & 75.24 & \textbf{52.00} & 60.00 & \textbf{87.50} & \textbf{87.50} \\
         & FT & 86.41 & 97.32 & \textbf{62.14} & \textbf{85.44} & \textbf{52.00} & \textbf{68.00} & 50.00 & 62.50 \\
        \midrule
        \multirow{4}{*}{17} & Linear & 82.00 & 93.60 & 46.34 & 73.17 & 34.29 & 54.29 & 50.00 & 50.00 \\
         & PCA & 82.40 & 93.47 & 46.83 & 71.71 & 34.29 & 57.14 & 50.00 & 50.00 \\
         & ALSPAC & \textbf{95.87} & \textbf{99.73} & 54.63 & 79.02 & 51.43 & 68.57 & 37.50 & \textbf{75.00} \\
         & Pre & 85.51 & 94.60 & 48.99 & 69.19 & 48.48 & 66.67 & 37.50 & 50.00 \\
         & FT & 93.81 & 98.95 & \textbf{68.18} & \textbf{85.35} & \textbf{60.61} & \textbf{72.73} & \textbf{62.50} & 62.50 \\
        \midrule
        \multirow{4}{*}{24} & Linear & 65.46 & 80.12 & 44.80 & 62.40 & 41.67 & 54.17 & 25.00 & 25.00 \\
         & PCA & 66.27 & 79.92 & 44.00 & 62.40 & 41.67 & 54.17 & 25.00 & 25.00 \\
         & ALSPAC & \textbf{95.78} & \textbf{99.60} & 56.80 & 80.00 & 37.50 & 62.50 & 75.00 & \textbf{87.50} \\
         & Pre & 85.48 & 95.09 & 52.86 & 73.57 & 50.00 & 61.11 & \textbf{87.50} & \textbf{87.50} \\
         & FT & 93.25 & 99.39 & \textbf{65.00} & \textbf{90.71} & \textbf{61.11} & \textbf{72.22} & \textbf{87.50} & \textbf{87.50} \\
        \bottomrule
    \end{tabular}
    \caption{{\bf Extrapolation.} Predicting the next masked time point (\%), all future points masked. Age 15 from \{9,11,13\}; 17 from \{9,11,13,15\}; 24 from \{9,11,13,15,17\}.
Methods: linear extrapolation (Linear), PCA extrapolation (PCA), ALSPAC-only
training (ALSPAC), pre-trained (Pre) and fine-tuned (FT), all on the ALSPAC
test set. Column headers are angular tolerances per severity band; best per
age/band/tolerance in bold.}
    \label{tab:res_n_1_two_tol}
\end{table}

\begin{figure}[h!]
    \centering
    \includegraphics[width=\linewidth]{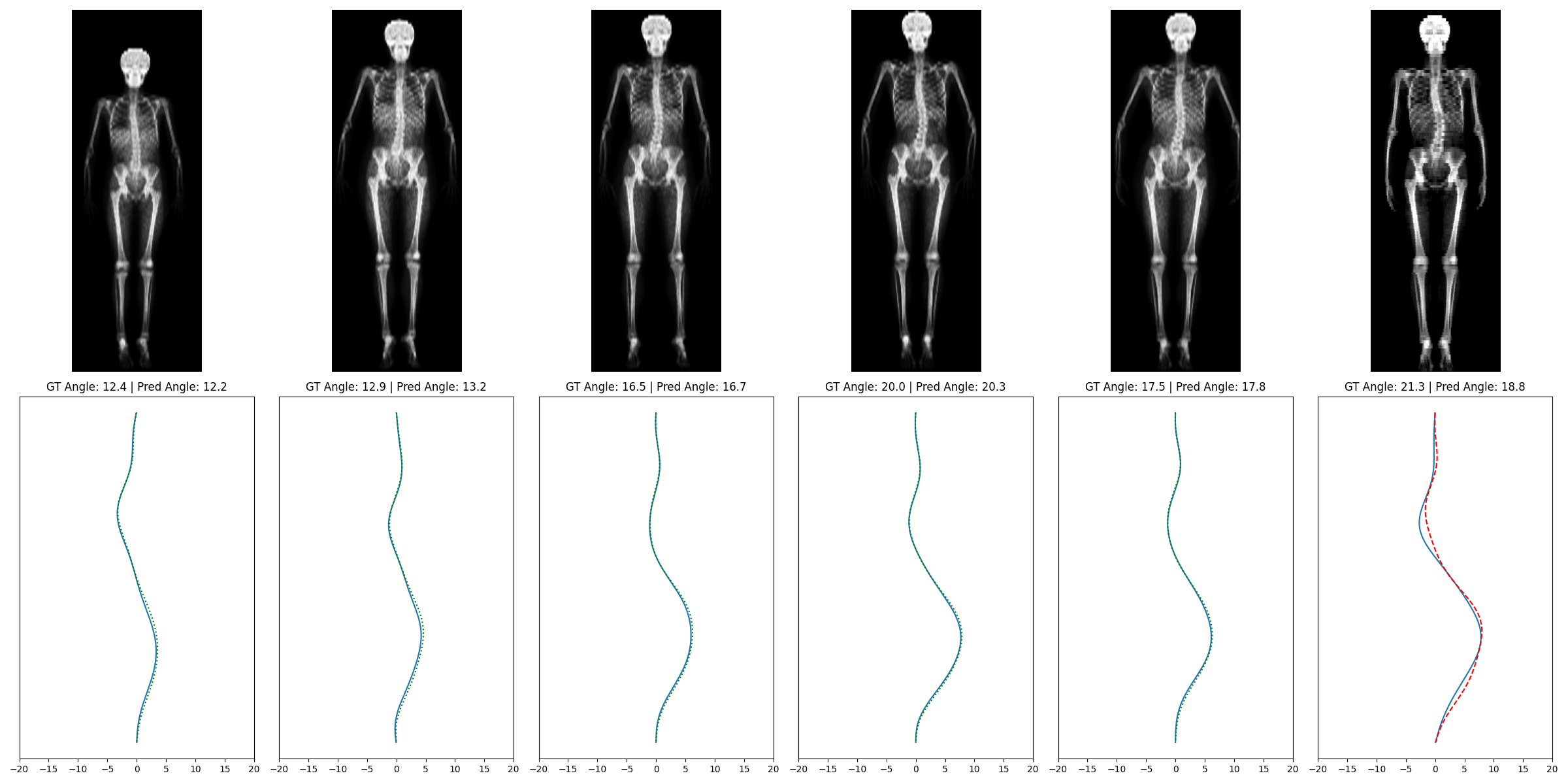}
    \caption{An extrapolation or next time point prediction (N+1) with 5 unmasked time points (green dotted) at 9, 11, 13, 15, 17 to predict time point 24 (red dashed). Time point 24 is masked to prevent the model from looking ahead at any future time points.}
    \label{fig:n_1_example_1}
\end{figure}

The task of predicting the next time point is important for clinical practice. The severity of Scoliosis can be measured automatically but for monitoring progression, it is preferable to know whether an individual is at risk of severe disease before they experience it. The next time point has been masked along with all subsequent time points, to ensure there is no visibility of the later time steps. 

The next time point is predicted  and measured against the ground truth. The model was pre-trained on synthetic data and has been evaluated on real data: after pre-training and again after fine-tuning, the results are  shown in table~\ref{tab:res_n_1_two_tol}. In the predicting the next timepoint task we show good results performing well against the baseline, our method shows more robust results, being able to predict complex patterns of increase, allowing for generalisable predictions. Predicting the next time point with good accuracy across a range of Scoliosis severities. Whilst occasionally getting good results in some bands, the PCA baseline struggles to preform reliably across all bands occasionally failing to estimate the next time point within a 5 degrees from the ground truth.

The model trained only on ALSPAC data, preforms well extrapolating 0-6\textdegree~angles but does not preform as well as the pre-trained models on the more complex moderate and severe cases, this is likely because severe scoliosis is rare in the real-world data and this model has struggled to learn these more complex increase patterns. The pre-trained model preforms well, fine-tuning  increases the accuracy of the results and also improves the consistency, generalisability and robustness of the model, improving the quality of predictions across ages 15, 17 and 24. An example of a next timepoint prediction is shown in 
Fig.~\ref{fig:n_1_example_1}.
\section{Conclusion}
We have shown that a transformer model pre-trained on synthetic data, and fine-tuned on real data can be used to predict spine curves in DXA scans of real Adolescent Idiopathic Scoliosis temporal sequences.

Uses of this model include as a potential future medical device, where the next time point could be predicted and patients could be followed up more frequently if they are likely to be at risk of progression to severe disease or given earlier access to physiotherapy and non-surgical bracing interventions before severe disease occurs. Another potential use is as an advanced analytics tool to help researchers impute missing observations in medical studies, allowing researchers without a machine learning background (e.g.\ epidemiologists) to impute missing time points with greater precision on the likely true value. Future work will include incorporating contextual information, such as: sex, height, weight \& skeletal maturity into predictions. As well as cross institute testing.\newline

\noindent{\bf Acknowledgements.} We are grateful to our funders: EPSRC CDT in Health Data Science (EP/S02428X/1), and the EPSRC programme grant Visual AI (EP/T025872/1). We also thank the ALSPAC study for providing us with access to the data. If others would like to use the data and/or annotations they can apply to the ALSPAC study to access this. Ethical approval for the study was obtained from the ALSPAC Ethics and Law Committee and the Local Research Ethics Committees. This study used data from the UK Biobank AUGMENT study, available to approved researchers through the standard UK Biobank access process (application number 17295).

\bibliographystyle{splncs04}
\bibliography{references/additional_refs,references/egbib}

@article{Li2024-cc,
  title    = "The application of machine learning methods for predicting the
              progression of adolescent idiopathic scoliosis: a systematic
              review",
  author   = "Li, Lening and Wong, Man-Sang",
  journal  = "BioMedical Engineering OnLine",
  volume   =  23,
  number   =  1,
  pages    = "80",
  month    =  aug,
  year     =  2024
}

@article{jamaludin_predicting_2019,
    title = {Predicting scoliosis in {DXA} scans using intermediate representations},
    volume = {11397 LNCS},
    issn = {16113349},
    url = {https://link.springer.com/chapter/10.1007/978-3-030-13736-6_2},
    doi = {10.1007/978-3-030-13736-6_2},
    journal = {Lecture Notes in Computer Science (including subseries Lecture Notes in Artificial Intelligence and Lecture Notes in Bioinformatics)},
    author = {Jamaludin, Amir and Kadir, Timor and Clark, Emma and Zisserman, Andrew},
    year = {2019},
    note = {ISBN: 9783030137359
Publisher: Springer Verlag},
    pages = {15--28},
}

@InProceedings{Pullen25,
  author       = "Owen Pullen and Amir Jamaludin and Andrew Zisserman",
  title        = "A Simple Modality-Agnostic Representation for Scoliosis Phenotyping",
  booktitle    = "MICCAI Workshop on Shape in Medical Imaging ",
  series       = "Lecture Notes in Computer Science",
  volume       = "16171",
  pages        = "204-217",
  month        = "sep",
  year         = "2025",
  publisher    = "Springer",
  doi          = "10.1007/978-3-032-06774-6",
}

@article{konieczny_epidemiology_2013,
    title = {Epidemiology of adolescent idiopathic scoliosis},
    volume = {7},
    issn = {18632548},
    url = {https://journals.sagepub.com/doi/10.1007/s11832-012-0457-4},
    doi = {10.1007/S11832-012-0457-4},
    number = {1},
    journal = {Journal of Children's Orthopaedics},
    author = {Konieczny, Markus Rafael and Senyurt, Hüsseyin and Krauspe, Rüdiger},
    month = feb,
    year = {2013},
    pmid = {24432052},
    note = {Publisher: Springer Verlag},
    pages = {3--9},
}

@article{boyd_cohort_2013,
    title = {Cohort {Profile}: {The} ‘{Children} of the 90s’—the index offspring of the {Avon} {Longitudinal} {Study} of {Parents} and {Children}},
    volume = {42},
    issn = {0300-5771},
    url = {https://dx.doi.org/10.1093/ije/dys064},
    doi = {10.1093/IJE/DYS064},
    number = {1},
    journal = {International Journal of Epidemiology},
    author = {Boyd, Andy and Golding, Jean and Macleod, John and Lawlor, Debbie A. and Fraser, Abigail and Henderson, John and Molloy, Lynn and Ness, Andy and Ring, Susan and Smith, George Davey},
    month = feb,
    year = {2013},
    pmid = {22507743},
    note = {Publisher: Oxford Academic},
    pages = {111--127},
}

@InProceedings{lin_2020_seg,
author="Lin, Yi
and Zhou, Hong-Yu
and Ma, Kai
and Yang, Xin
and Zheng, Yefeng",
editor="Cai, Yunliang
and Wang, Liansheng
and Audette, Michel
and Zheng, Guoyan
and Li, Shuo",
title="Seg4Reg Networks for Automated Spinal Curvature Estimation",
booktitle="Computational Methods and Clinical Applications for Spine Imaging",
year="2020",
publisher="Springer International Publishing",
address="Cham",
pages="69--74",
isbn="978-3-030-39752-4"
}

@InProceedings{lin_2021_seg,
author="Lin, Yi
and Liu, Luyan
and Ma, Kai
and Zheng, Yefeng",
editor="de Bruijne, Marleen
and Cattin, Philippe C.
and Cotin, St{\'e}phane
and Padoy, Nicolas
and Speidel, Stefanie
and Zheng, Yefeng
and Essert, Caroline",
title="Seg4Reg+: Consistency Learning Between Spine Segmentation and Cobb Angle Regression",
booktitle="Medical Image Computing and Computer Assisted Intervention -- MICCAI 2021",
year="2021",
publisher="Springer International Publishing",
address="Cham",
pages="490--499",
isbn="978-3-030-87240-3"
}

@article{ZHANG2022sco,
title = {MPF-net: An effective framework for automated cobb angle estimation},
journal = {Medical Image Analysis},
volume = {75},
pages = {102277},
year = {2022},
issn = {1361-8415},
doi = {10.1016/j.media.2021.102277},
url = {https://www.sciencedirect.com/science/article/pii/S1361841521003224},
author = {Kailai Zhang and Nanfang Xu and Chenyi Guo and Ji Wu}
}

@misc{radiopaedia_cobb,
  author={Thuaimer, A. and Knipe, H. and Elfeky, M.},
  title={Cobb angle},
  year={2021},
  howpublished={Radiopaedia.org, reference article},
  doi={10.53347/rID-23612},
  url={https://radiopaedia.org/articles/cobb-angle},
}

@article{cobb1948outline,
  title={Outline for the study of scoliosis},
  author={Cobb, JR},
  journal={Instr Course Lect AAOS},
  volume={5},
  pages={261--275},
  year={1948}
}

@misc{jamaludin2023predicting,
      title={Predicting Spine Geometry and Scoliosis from DXA Scans}, 
      author={Amir Jamaludin and Timor Kadir and Emma Clark and Andrew Zisserman},
      year={2023},
      eprint={2311.09424},
      archivePrefix={arXiv},
      primaryClass={cs.CV},
      url={https://arxiv.org/abs/2311.09424}, 
}

@article{devlin2018bert,
  title={Bert: Pre-training of deep bidirectional transformers for language understanding},
  author={Devlin, Jacob and Chang, Ming-Wei and Lee, Kenton and Toutanova, Kristina},
  journal={arXiv preprint arXiv:1810.04805},
  year={2018}
}

@inproceedings{
bourigault2022scoliosis,
title={Scoliosis Measurement on {DXA} Scans Using a Combined Deep Learning and Spinal Geometry Approach},
author={Emmanuelle Bourigault and Amir Jamaludin and Timor Kadir and Andrew Zisserman},
booktitle={MIDL},
year={2022},
}

@InProceedings{wu2017automatic,
author="Wu, Hongbo
and Bailey, Chris
and Rasoulinejad, Parham
and Li, Shuo",
title="Automatic Landmark Estimation for Adolescent Idiopathic Scoliosis Assessment Using BoostNet",
booktitle="MICCAI",
year="2017",
}

@article{taylor_identifying_2013,
    title = {Identifying scoliosis in population-based cohorts: {Development} and validation of a novel method based on total-body dual-energy {X}-ray absorptiometric scans},
    volume = {92},
    issn = {0171967X},
    url = {https://link.springer.com/article/10.1007/s00223-013-9713-y},
    doi = {10.1007/S00223-013-9713-Y},
    number = {6},
    journal = {Calcified Tissue International},
    author = {Taylor, Hilary J. and Harding, Ian and Hutchinson, John and Nelson, Ian and Blom, Ashley and Tobias, Jon H. and Clark, Emma M.},
    month = jun,
    year = {2013},
    pmid = {23456028},
    note = {Publisher: Springer},
    pages = {539--547},
}

@misc{miccai_aasce_2019,
    title = {{AASCE} - {MICCAI} 2019 {Challenge}: {Accurate} {Automated} {Spinal} {Curvature} {Estimation}},
    url = {https://aasce19.github.io/},
    author = {MICCAI},
    year = {2019},
}
\end{document}